\documentclass[english]{IEEEtran}
\usepackage[T1]{fontenc}
\usepackage{textcomp}
\UseRawInputEncoding
\usepackage{babel}
\usepackage{array}
\usepackage{booktabs}
\usepackage{multirow}
\usepackage{amsbsy}
\usepackage{amstext}
\usepackage{amssymb}
\usepackage{graphicx}
\usepackage[bookmarks=false,
 breaklinks=false,pdfborder={0 0 1},backref=page,colorlinks=false]
 {hyperref}

\makeatletter

\providecommand{\tabularnewline}{\\}

\usepackage[noadjust]{cite}

\AtBeginDocument{
  
}

\makeatother

\begin{document}
\title{Task-Oriented Grasping Using Reinforcement Learning with a Contextual
Reward Machine}
\author{Hui~Li, Akhlak Uz Zaman, Fujian Yan, and~Hongsheng~He \thanks{Hui Li, Akhlak Uz Zaman and Hongsheng He are with the department of
Computer Science, The University of Alabama, Tuscaloosa, AL, 35487,
USA. This research is funded by NSF \#2420355.}\thanks{Fujian Yan is with the School of Computing, Wichita State University,
Wichita, KS, 67260, USA.}\thanks{Correspondence should be addressed to Hongsheng He, e-mail: \protect\href{http://hongsheng.he@ua.edu}{hongsheng.he@ua.edu}. }\thanks{\textbf{This work has been submitted to the IEEE for possible publication.
Copyright may be transferred without notice, after which this version
may no longer be accessible.}}}
\maketitle
\begin{abstract}
This paper presents a reinforcement learning framework that incorporates
a Contextual Reward Machine for task-oriented grasping. The Contextual
Reward Machine reduces task complexity by decomposing grasping tasks
into manageable sub-tasks. Each sub-task is associated with a stage-specific
context, including a reward function, an action space, and a state
abstraction function. This contextual information enables efficient
intra-stage guidance and improves learning efficiency by reducing
the state-action space and guiding exploration within clearly defined
boundaries. In addition, transition rewards are introduced to encourage
or penalize transitions between stages which guides the model toward
desirable stage sequences and further accelerates convergence. When
integrated with the Proximal Policy Optimization algorithm, the proposed
method achieved a 95\% success rate across 1,000 simulated grasping
tasks encompassing diverse objects, affordances, and grasp topologies.
It outperformed the state-of-the-art methods in both learning speed
and success rate. The approach was transferred to a real robot, where
it achieved a success rate of 83.3\% in 60 grasping tasks over six
affordances. These experimental results demonstrate superior accuracy,
data efficiency, and learning efficiency. They underscore the model\textquoteright s
potential to advance task-oriented grasping in both simulated and
real-world settings.
\end{abstract}


\section{Introduction}

Robotic dexterity, the ability of a robot to manipulate objects with
precision, adaptability, and control, akin to human hand dexterity,
is essential for performing complex tasks across diverse applications,
including aerospace, automotive, manufacturing, and warehousing, and
medical rehabilitation \cite{fan2017robust,chen2025target}. While
robots excel in structured environments and repetitive tasks, they
remain constrained in unstructured and dynamic scenarios. Advancing
robotic dexterity has the potential to bridge this gap and allow robots
to handle complex tasks in uncertain environments \cite{james2020slip}.
The current methods struggle with dexterous manipulation, especially
when handling objects of varying shapes, sizes, and materials.

Grasping an object represents the initial and foundational step of
dexterous manipulation. A key factor in grasping tasks is the selection
of an appropriate grasp topology, which defines the specific configuration
of a robotic hand when interacting with an object. The grasp topology
plays a pivotal role in minimizing redundant finger movements and
streamlining the overall manipulation process. Selecting a suitable
topology is a critical prerequisite for effective manipulation, as
it ensures stable object acquisition and simplifies downstream control
\cite{wang2023dexgraspnet}. A firm or adaptive grasp stabilizes the
object and enhances task efficiency and execution success.

Research in this area has explored various strategies for determining
effective grasp topologies, including learning from human demonstrations
to replicate natural grasp patterns \cite{chen2025towards,mandikal2022dexvip},
designing soft robotic hands for flexible and compliant interactions
\cite{deimel2016novel}, and leveraging advanced learning-based techniques
for data-driven optimization \cite{duan2021robotics}. Selecting a
grasp topology that aligns with object properties and task requirements
reduces manipulation complexity and improves performance across diverse
scenarios \cite{shang2020deep}.

While numerous grasp types exist, they generally cluster into a limited
set that is suitable for manipulating everyday objects and tools \cite{feix2016grasp,santello1998postural}.
This insight has led to the development of grasp taxonomy that categorizes
and simplifies grasp poses \cite{cutkosky1989grasp,bullock2015yale}.
The taxonomy provides a systematic way to map object characteristics
and task requirements to suitable grasp types, thereby improving planning
and control.

The choice of grasp topology is influenced by both object features,
such as size, shape, surface texture, and mass, as well as task-specific
requirements, including the intended action, applied forces, and environmental
constraints \cite{rao2019object,feix2014analysis}. For instance,
precision tasks, like picking up small or fragile objects, typically
utilize pinch or tripod grasps. In contrast, tasks requiring greater
stability or force, such as lifting heavy items, benefit from power
grasps. Tool-use scenarios introduce additional complexity and often
necessitate specialized topologies, such as cylindrical grasps for
tool handles or lateral grasps for flat items. Building on these insights,
our previous work \cite{li2021learning} successfully integrated object
features and task requirements into grasp strategies and established
a structured framework for the grasping process.

\begin{figure*}[t]
\centering{}\includegraphics[width=0.98\textwidth]{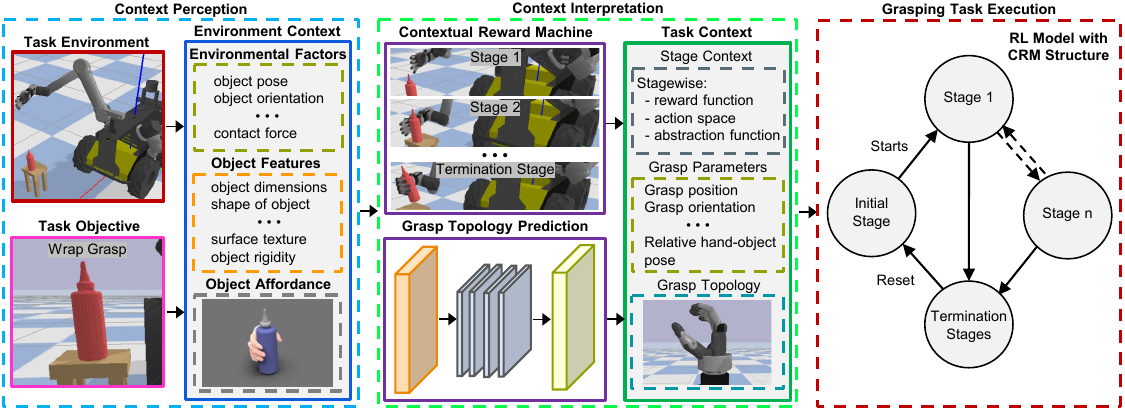}\caption{Context-aware task-oriented grasping framework with a contextual reward
machine. \label{fig: task_framework}}
\end{figure*}

Deploying grasping tasks presents significant challenges due to environmental
uncertainties, particularly in unstructured environments. The perception
of target objects is often incomplete or inaccurate, which negatively
impacts grasping performance. Traditional planning methods struggle
to manage these complexities because they rely on precise context
modeling, which is rarely feasible in real-world scenarios.

Task-oriented grasping can be formulated as a sequential decision-making
problem in which an agent learns optimal behaviors through trial and
error to maximize cumulative rewards. Reinforcement learning (RL)
has shown strong potential in addressing such problems effectively
\cite{sutton2018reinforcement,qin2023dexpoint,mandikal2021learning}.

RL has demonstrated particular success in solving complex control
problems in robotics, especially when integrated with Proximal Policy
Optimization (PPO) \cite{thuruthel2018model,stulp2012reinforcement,he2020reinforcement}.
PPO, a widely used RL algorithm, addresses many limitations of RL
by improving training stability and sample efficiency \cite{schulman2017proximal}.
Its simplicity and robustness make it well suited for high-dimensional
and continuous action spaces, which enable effective policy optimization.

Despite these advantages, RL methods for grasping still face several
challenges, including high computational demands and difficulties
in achieving stable convergence. These limitations highlight the need
for further development of RL techniques that can meet the demands
of real-world applications such as agriculture, communication and
robotics \cite{li2022reinforcement,li2021q}.

To address these challenges, researchers have proposed multistage
reinforcement learning approaches where each stage of a task is trained
separately using specialized sub-networks that collaborate to determine
an overall optimal policy \cite{yang2019deep}. Although this method
achieves stable convergence, it remains computationally expensive.
The reward machine framework has been introduced to solve the problems
by organizing complex tasks into modular sub-tasks, each associated
with a distinct reward function. This structure improves learning
efficiency and reduces computation cost \cite{icarte2022reward}.
Although reward machines offer a structured approach to task decomposition,
traditional implementations often lack the flexibility to adapt to
dynamic environments or incorporate detailed contextual information.
These constraints limit their applicability to real-world scenarios,
where adaptability and context-awareness are essential for reliable
robotic performance.

In this paper, we propose a context-aware task-oriented grasping approach
that leverages a Contextual Reward Machine (CRM) to enhance efficiency
and adaptability. The CRM decomposes grasping tasks into sequential
stages with each stage defined by a stage-specific context including
a reward function, an action space, and abstracted states. This structure
guides intra-stage task progression and improves learning efficiency.
Additionally, a transition reward mechanism is designed to facilitate
smooth transitions between stages.

The general structure of the proposed method is illustrated in Fig.
\ref{fig: task_framework}. In this approach, the context of the environment
for a grasping task is continuously perceived and analyzed. The environmental
context comprises object features, such as dimensions, shape, and
texture. It also includes environmental factors, such as object pose,
contact forces, obstacle positions, and object affordances. The object
features and the task objective are processed by a pretrained grasp
selection network to determine the appropriate grasp topology. Meanwhile,
the environmental factors and object affordances are used by the CRM
to identify the grasp location, determine the current stage, and retrieve
the corresponding stage-specific context. They are also utilized by
the RL agent to learn and optimize its policy. Task execution is carried
out by an RL model, which integrates the grasp topology, grasp location,
and stage-specific contexts under CRM framework. The model dynamically
adapts to the changing conditions and performs robust, precise, and
efficient grasps.

The main contributions of the paper include:
\begin{itemize}
\item We implemented a context-aware task-oriented dexterous grasping approach
that enables adaptive and efficient grasping in unstructured environments.
\item We designed and developed a Contextual Reward Machine that decomposes
grasping tasks into sequential stages. It utilizes stage-specific
contexts and transition rewards to enhance learning efficiency and
adaptability.
\end{itemize}

\section{Framework of Task-Oriented Grasping}

Task-oriented grasping is inherently a context-aware process involving
the perception of environmental context, the generation of grasp strategies
based on that context, and the efficient, adaptive execution of those
strategies. To address these challenges, environmental context was
obtained through sensor fusion by integrating inputs from multiple
sensors. A deep learning network was developed to generate grasp strategies,
while a reinforcement learning model under CRM framework was designed
and implemented to enable efficient and adaptive execution of grasping
tasks.

\subsection{Context Perception}

Task-oriented grasping tasks require detailed environmental context.
An RGB-D camera is used to capture object dimensions, shape, and surface
characteristics. These object features guide the selection of an appropriate
grasp topology. Environmental factors, such as the object's position,
orientation, and nearby obstacles, are also captured with the same
RGB-D camera. In parallel, force-sensing resistors (FSRs) are used
to record contact forces and provide precise pressure data to improve
task accuracy.

\subsection{Grasp Topology Determination}

A proper grasp topology facilitates smoother and more efficient manipulation.
To simplify the selection process, we defined a grasp taxonomy comprising
six primary topologies and developed a grasp selection network \cite{li2021learning}
that maps object features and task demands to the most suitable grasp
topology. 
\begin{figure}[h]
\centering{}\includegraphics[width=0.98\columnwidth]{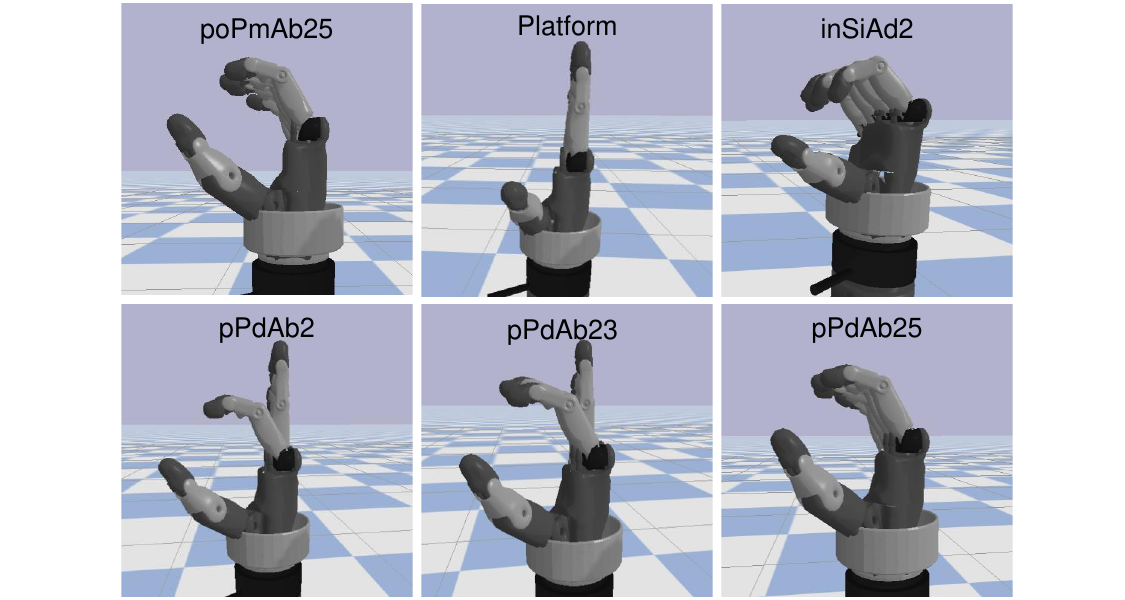}\caption{Grasp taxonomy with six grasp topology: the grasp topology names represent
grasp attributes where \textquotedbl In,\textquotedbl{} \textquotedbl po,\textquotedbl{}
and \textquotedbl p\textquotedbl{} indicate intermediate, power,
and precision grasps. \textquotedbl Si,\textquotedbl{} \textquotedbl Pm,\textquotedbl{}
and \textquotedbl Pd\textquotedbl{} refer to side, palm, and pad
opposition. \textquotedbl Ab\textquotedbl{} and \textquotedbl Ad\textquotedbl{}
signify abduction and adduction, and numbers define virtual finger
groups. \label{fig: topology}}
\end{figure}

The adopted taxonomy is illustrated in Fig. \ref{fig: topology}:
(1) the platform grasp for holding, pushing, or pressing; (2) the
power grasp (poPmAb25) for securely gripping objects; (3) precision
grasps (pPdAb2, pPdAb23, pPdAb25) for tasks requiring fine dexterity;
and (4) the intermediate grasp (InSiAd2) for levering or twisting
actions. 

The grasp selection network is a multi-class, multi-label Multilayer
Perceptron (MLP) neural network that takes object features and task
objectives as input and predicts the probability of each grasp topology
in the predefined taxonomy. The topology with the highest probability
is selected as the target grasp pose.

\subsection{Execution of Grasping Tasks}

\begin{figure}
\centering{}\includegraphics[width=0.98\columnwidth]{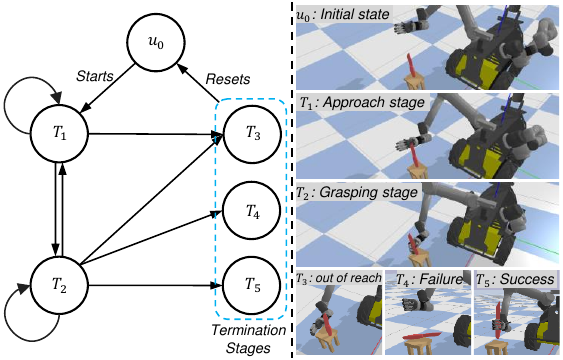}

\caption{A general example of grasping task decomposition. \label{fig:  states_env}}
\end{figure}

We used a reinforcement learning approach for grasping task execution.
Due to the inherent complexity of grasping, such tasks often exhibit
limited flexibility and present optimization challenges. To reduce
task complexity, we decomposed the grasping process into a sequence
of manageable stages. Although grasping tasks can be decomposed in
various ways, Fig. \ref{fig:  states_env} shows a general decomposition
framework. In the framework, the initial state represents the initial
configuration of the task environment. In the\textbf{ }approach stage,
the robot hand moves to an appropriate position and orientation (grasp
location) for grasping. During the grasping stage, the fingers adjust
to establish a stable or adaptive grasp based on the task's requirements.
Finally, the termination stages reflect task outcomes, such as grasp
success, grasp failure, or the object being out of reach, and signify
the completion of the task. 

This method requires stage-specific learning mechanisms as each stage
operates within a distinct context. To address this problem, we propose
a Contextual Reward Machine that explicitly defines and manages these
contexts.

\section{Contextual Reward Machine}

The CRM provides a structured and interpretable framework for addressing
complex tasks by encoding task-specific knowledge into a hierarchical
representation. This structure enables adaptive rewards based on task
progress and stage transitions toward the desired goal. By guiding
the agent through task-relevant stages, this approach improves learning
efficiency and supports effective, goal-directed behaviors. Further
details are provided in this section.

\subsection{The General Framework of CRM}

The general framework of the CRM extends the standard reward machine
\cite{icarte2022reward} by incorporating task context and a stage
transition function, formally defined as
\begin{equation}
\mathcal{M}=(U,u_{0},\Sigma,\delta,\mathcal{T},R_{\text{T}})\label{eq: tsrm_framework}
\end{equation}
where $\mathcal{T}=\{(A_{i},r^{(i)},\phi_{i})\}_{i=1}^{k}$ represents
a set of $k$ stages (or sub-tasks), each characterized by task-specific
knowledge. Each stage $T_{i}$ consists of an action set $A_{i}\subseteq\mathcal{A}$,
a state abstraction function $\phi_{i}:U\to U_{i}'$, which maps the
global state space $U$ to a stage-relevant abstract state space $U_{i}'$
to simplify stage representation, and a stage reward function $r^{(i)}:U_{i}'\times A_{i}\to\mathbb{R}$
which defines the rewards for actions within the stage $T_{i}$. 

The transition function $\delta:U\times\Sigma\to\mathcal{T}$ determines
whether a stage transition occurs and, if so, to which stage, based
on the current state and $\Sigma$, the set of events triggering stage
transitions. A transition occurs when the current state satisfies
one of these events. Upon a stage transition, the stage transition
history set $\mathcal{H}\subseteq\mathcal{T}\times\mathcal{T}$ is
updated as 
\begin{equation}
\mathcal{H}=\{(T_{i},T_{j})\mid i,j=1,2,\dots,k\}
\end{equation}
where each transition $(T_{i},T_{j})\in\mathcal{H}$ is associated
with a reward $R_{\text{T}}:\mathcal{H}\to\mathbb{R}$, which quantifies
the desirability of the transition. 
\begin{figure}
\centering{}\includegraphics[width=0.96\columnwidth]{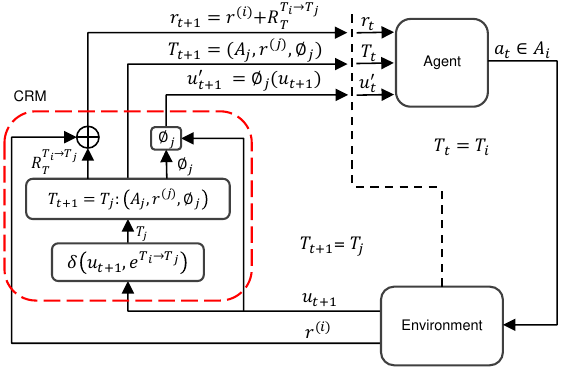}

\caption{The Contextual Reward Machine: The dotted line separates the RL processes
at timesteps $t$ and $t+1$, illustrating the sequential interaction
between the agent, environment, and the CRM. \label{fig: RL_CRM}}
\end{figure}

The CRM within the RL framework operates iteratively and allows the
agent to interact with the environment and adapt dynamically, as illustrated
in Fig. \ref{fig: RL_CRM}. At timestep $t$, the agent operates within
stage $T_{t}=T_{i}$. Based on the abstract state $u'_{t}$ and action
space $A_{i}$, the agent selects an action $a_{t}\in A_{i}$. The
environment responds with the next global state $u_{t+1}$ and an
intra-stage reward $r^{(i)}(u'_{t},a_{t})$ which reflects the action's
outcome.

The CRM processes the feedback to determine the next stage $T_{t+1}=T_{j}$
using the transition function $\delta(u_{t+1},e^{T_{i}\rightarrow T_{j}})$,
where $e^{T_{i}\rightarrow T_{j}}\in\Sigma$ represents the triggering
event. If no transition occurs $(i=j)$, the global state $u_{t+1}$
is abstracted to $u'_{t+1}$ using $\phi_{i}$. If a transition occurs
$(i\neq j)$, the global state $u_{t+1}$ is abstracted to the corresponding
state $u'_{t+1}$ of stage $T_{j}$ using $\phi_{j}$. These abstraction
functions extract task-relevant features and simplify state representation.

The reward at timestep $t+1$ is computed as 
\begin{equation}
r_{t+1}=r^{(i)}(u'_{t},a_{t})+R_{\text{T}}^{T_{i}\rightarrow T_{j}}
\end{equation}
and the cumulative reward is expressed as 
\begin{equation}
R=\sum_{t=1}^{T}r_{t}(u'_{t},a_{t})+\sum_{(i,j)\in\mathcal{H}}R_{\text{T}}^{T_{i}\rightarrow T_{j}}
\end{equation}
where $r_{t}$ represents the stage-specific reward at timestep $t\in[1,\dots,T]$,
and $T$ is the total number of timesteps. The first term captures
intra-stage rewards, while the second accounts for transition rewards.
The stage knowledge in $T_{t+1}$, the abstract state $u'_{t+1}$,
and the reward $r_{t+1}$ define the context for the next timestep
which supports efficient task decomposition and reward optimization
for solving complex tasks. 

\subsection{PPO with CRM}

To adapt PPO to the CRM framework, the objective function is revised
to align with the hierarchical structure of CRM. For a stage $T_{i}$
at timestep $t$, the clipped surrogate objective is
\begin{equation}
\begin{array}{cc}
\mathcal{L}_{\mathcal{M}}^{\text{CLIP}}(\theta)= & \mathbb{E}_{(u',a)\sim\pi_{\text{old}}}\Big[\min\big(r_{t}^{\text{\ensuremath{\mathcal{M}}}}(\theta)A_{t}^{\text{\ensuremath{\mathcal{M}}}},\\
 & \text{clip}(r_{t}^{\text{\ensuremath{\mathcal{M}}}},1-\epsilon,1+\epsilon)A_{t}^{\text{\ensuremath{\mathcal{M}}}}\big)\Big]
\end{array}
\end{equation}
where $r_{t}^{\text{\ensuremath{\mathcal{M}}}}(\theta)=\frac{\pi_{\theta}(a_{t}|u'_{t})}{\pi_{\text{old}}(a_{t}|u'_{t})}$
represents the probability ratio between the current and old policies
based on the abstract state $u'_{t}=\phi_{i}(u_{t})$. 

The advantage function $A_{t}^{\text{\ensuremath{\mathcal{M}}}}$
is defined as
\begin{equation}
A_{t}^{\text{\ensuremath{\mathcal{M}}}}=r_{t}(u'_{t},a_{t})+\gamma V(u'_{t+1})+R_{\text{T}}^{T_{i}\rightarrow T_{j}}-V(u'_{t})
\end{equation}
where $r_{t}(u'_{t},a_{t})=r^{(i)}(u'_{t},a_{t})$ is the intra-stage
reward, $R_{\text{T}}^{T_{i}\rightarrow T_{j}}$ is the transition
reward, and $V(u'_{t})$ and $V(u'_{t+1})$ are value estimates for
the current and next stages, respectively.

\section{CRM-PPO for Grasping Tasks}

The CRM-PPO model divides grasping tasks into distinct stages, each
defined by a specific context and transition mechanism. To optimize
task performance, it is crucial to clearly specify these contexts
and mechanisms for each stage. This section outlines the stage contexts
and corresponding transition mechanisms for each stage of the grasping
task.

\subsection{Task Decomposition }

We followed the task decomposition strategy shown in Fig. \ref{fig:  states_env}
to decompose grasping tasks using the general CRM framework (Eq. \ref{eq: tsrm_framework}).
The global state $U$ is defined as the set of all possible states
in the grasping environment, while $u_{\mathrm{initial}}=u_{0}$ represents
the initial state corresponding to the environment's default configuration.
Each grasping task begins from the initial state $u_{0}$, and upon
task completion, the environment resets to $u_{0}$ to prepare for
the next task.

The system transitions directly from the initial state to the approach
stage without receiving any reward. During the approach stage, the
robot moves its hand toward the object. If the object becomes out
of reach in this stage $(\ensuremath{e^{\mathrm{aor}}})$, the task
transitions to the out-of-reach stage with a penalty $R_{\text{T}}^{\mathrm{aor}}$.
Arriving at the grasp location $(\ensuremath{e^{\mathrm{arrive}}})$
transitions the task to the grasping stage, with a transition reward
$R_{\text{T}}^{\mathrm{\mathrm{arrive}}}$. The approach stage cannot
transition directly to the grasp-failure or grasp-success stages,
as a failed or successful grasp requires hand-object interaction,
which occurs only in the grasping stage.

In the grasping stage, the robot manipulates its fingers to attempt
a grasp. Knocking the object out of reach in the grasping stage $(\ensuremath{e^{\mathrm{gor}}})$
results in a transition to the out-of-reach stage with a penalty $R_{\text{T}}^{\mathrm{gor}}$.
Failure to grasp the object $(\ensuremath{e^{\mathrm{fail}}})$ results
in a transition to the grasp-failure stage with a penalty $R_{\text{T}}^{\mathrm{fail}}$.
Successfully grasping the object $(\ensuremath{e^{\mathrm{succ}}})$
leads to the grasp-success stage with a reward $R_{\text{T}}^{\mathrm{succ}}$.

The out-of-reach, grasp-success, and grasp-failure stages serve as
termination stages, which conclude the current task and reset the
environment to the initial state $u_{0}$ in preparation for the next
one.

\subsection{Action Space}

The designed stages of a task-oriented grasping include the approach
stage, the grasping stage, and the termination stages.

\subsubsection{The Approach Stage}

In the approach stage, the objective is to move the hand toward the
grasp location while avoiding collisions with the object. Finger movements
are disabled in the approach stage to reduce collision risks and improve
sample efficiency. The action space is defined as
\[
A_{\mathrm{approach}}=[\Delta x,\Delta y,\Delta z]
\]
where $\Delta x$, $\Delta y$, and $\Delta z$ represent incremental
changes along the $x$, $y$, and $z$ axes, respectively.

\subsubsection{The Grasping Stage}

The action space for the grasping stage is defined as
\[
A_{\mathrm{\mathrm{grasp}}}=[\Delta x,\Delta y,\Delta z,\theta_{\text{thumb}},\theta_{\text{index}},\theta_{\text{middle}},\theta_{\text{ring}},\theta_{\text{little}}]
\]
where $\theta_{\text{thumb}}$, $\theta_{\text{index}}$, $\theta_{\text{middle}}$,
$\theta_{\text{ring}}$, and $\theta_{\text{little}}$ represent the
proximal interphalangeal (PIP) joint angles of the thumb, index, middle,
ring, and little fingers, respectively. In the grasping stage, fine
adjustments to the hand position are crucial for achieving an optimal
grasp, even when the hand is close to the grasp location. To enable
these adjustments, the hand's movements remain constrained by $\Delta x$,
$\Delta y$, and $\Delta z$, but with reduced step sizes for finer
control. 

To enhance grasp quality, we developed a simplified hand model inspired
by human hand kinematics and anatomy. The model captures key biomechanical
features such as joint articulation and finger linkage and enables
more natural and effective grasping strategies. This model focuses
on finger flexion and extension while excluding finger spreading.
Joint angles are constrained based on the PIP joint angle, $\theta_{\text{PIP}}$,
with the following relationships $\theta_{\text{DIP}}=\alpha_{\text{DIP}}\cdot\theta_{\text{PIP}}$
and $\theta_{\text{MCP}}=\alpha_{\text{MCP}}\cdot\theta_{\text{PIP}}$
where $\theta_{\text{DIP}}$ and $\theta_{\text{MCP}}$ represent
the joint angles of the distal interphalangeal (DIP) joint and the
metacarpophalangeal (MCP) joint, respectively. The constants $\alpha_{\text{DIP}}$
and $\alpha_{\text{MCP}}$ define proportional joint coordination.
For the thumb, the relationship between the interphalangeal (IP) joint
angle $\theta_{\text{IP}}$ and the trapeziometacarpal (TMCP) joint
angle $\theta_{\text{TMCP}}$ is given by $\theta_{\text{IP}}=\alpha_{\text{TMCP}}\cdot\theta_{\text{TMCP}}$.
Here, the constant $\alpha_{\text{TMCP}}$ defines the proportional
coupling between the two joints, it reflects the biomechanical constraints
of human thumb movement. The corresponding values for $\alpha_{\text{DIP}}$
are 0.77, 0.75, 0.75, and 0.57 for the index, middle, ring, and little
fingers, respectively. The value of $\alpha_{\text{MCP}}$ is 0.67
for all fingers, while $\alpha_{\text{TMCP}}=0.5$, as reported in
\cite{mentzel2011dynamics,roda2022studying,hrabia2013whole}.

The out-of-reach, grasp-success, and grasp-failure stages are termination
stages and therefore have no associated action sets.

\subsection{Observation Space}

The observation space is defined as
\begin{equation}
\mathcal{O}=[n_{\text{c}},o_{\textnormal{dist}},\boldsymbol{o}_{\textnormal{object}},o_{\textnormal{cone}},\boldsymbol{o}_{\mathrm{\text{relative}}},\boldsymbol{o}_{\mathrm{\text{force}}},\boldsymbol{o}_{\mathrm{\text{torque}}}]
\end{equation}
where each component represents a critical aspect of the grasping
task. The variable $n_{\text{c}}$ indicates the number of contact
points between the robot hand and the object which reflects contact
extent and contributes to grasp stability. The distance $o_{\textnormal{dist}}$
measures the proximity of the robot hand to the grasp location. The
vector $\boldsymbol{o}_{\textnormal{object}}$ represents the object
position, which supports object out-of-range detection and task success
or failure evaluation.

The vector $\boldsymbol{o}_{\mathrm{relative}}$ describes the spatial
relationship between the robot hand and the object which provides
essential grasp configuration details. The vectors $\boldsymbol{o}_{\mathrm{force}}$
and $\boldsymbol{o}_{\mathrm{torque}}$ represent the summed magnitudes
of contact forces and torques at all contact points along the $x$,
$y$, and $z$ axes. They enable the evaluation of force and torque
equilibrium. Together, these components comprehensively describe the
grasping task which covers grasp position, configuration, and stability. 

The Boolean variable $o_{\text{cone}}$ indicates whether all contact
forces lie within the friction cone and ensures stability through
frictional constraints. The friction cone is defined by the coefficient
of friction $\mu$ and the normal force $\mathbf{F}_{n}$, which satisfies
the condition: $\|\mathbf{F}_{c}\|\leq\mu\cdot\mathbf{F}_{n}$ where
$\mathbf{F}_{c}$ is the contact force vector. Grasp stability is
determined as $o_{\text{cone}}=1$ if the above condition is satisfied
for all contact points, and $o_{\text{cone}}=0$ otherwise. This ensures
that all contact forces remain within the friction cone which prevents
slippage and enhances grasp stability. 

Each stage of the grasping task has specific goals and therefore requires
specific information. The state abstraction function $\phi_{i}$ extracts
the task-relevant information from the observation space which reduces
computational complexity and improves efficiency.

\subsection{Reward Function}

\subsubsection{The Approach Stage}

The objective of this stage is to move the robot hand as close as
possible to the grasp location while avoiding collisions and ensuring
the object remains within the workspace. The reward function for this
stage is defined as
\begin{equation}
r_{\text{appr}}=r^{\text{dist}}-\rho_{\mathrm{appr}}n_{\text{c}}
\end{equation}
where $\rho_{\mathrm{appr}}$ is a constant coefficient, and the distance-based
reward is defined as $r^{\text{dist}}=-\mathrm{e}^{|o_{\text{dist}}|}$,
which is inversely proportional to the distance between the hand and
the grasp location. The exponential form ensures rapid changes when
the hand is far from the target and more gradual changes as it approaches
the object. It encourages larger adjustments at greater distances
and finer movements when closer. This enhances both the efficiency
and accuracy of the task.

To reduce the risk of unintended collisions, which could cause the
object to be knocked out of the workspace, a penalty proportional
to the number of contact points $\ensuremath{n_{\text{c}}}$ is applied.
This reward design encourages precise and collision-free hand movements
during the approach stage. 

\subsubsection{The Grasping Stage}

The reward function for the grasping stage is designed to encourage
stable and efficient grasping behaviors. The model is rewarded based
on the number of contact points $n_{\text{c}}$, as a higher number
of contact points leads to increased grasp stability. Additionally,
the reward function evaluates equilibrium by minimizing net forces
$\boldsymbol{o}_{\text{force}}$ and torques $\boldsymbol{o}_{\text{torque}}$
along the three axes at all contact points. The closer these values
are to zero, the higher the reward, which indicates a more stable
and secure grip. The reward function for the grasping stage is defined
as
\begin{equation}
r_{\mathrm{grasp}}=r^{\text{equil}}+\rho_{\mathrm{grasp}}n_{\text{c}}+R_{\text{T}}^{\mathrm{arrive}}
\end{equation}
where the equilibrium reward is expressed as
\begin{equation}
r^{\text{equil}}=-\mathrm{e}^{\mid\boldsymbol{o}_{\text{force}}\mid}-\mathrm{e}^{\mid\boldsymbol{o}_{\text{torque}}\mid}
\end{equation}
The coefficient $\rho_{\mathrm{grasp}}$ adjusts the reward\textquoteright s
sensitivity to the number of contact points. The stage transition
reward $R_{\text{T}}^{\mathrm{arrive}}=R_{\text{T}}^{T_{1}\rightarrow T_{2}}$
encourages the transition from the approach stage to the grasping
stage. Its value is set higher than the maximum achievable reward
in the approach stage $r_{\text{appr}}$ to ensure that transitioning
to the grasping stage is prioritized, while still maintaining a stable
and controlled grip.

\subsubsection{The Termination Stages}

The out-of-reach, the grasp-success, and the grasp-failure stages
monitor task outcomes, where the reward is only related to the position
of the object $\boldsymbol{o}_{\mathrm{\text{object}}}$. The reward
function for the out-of-reach stage is defined as
\begin{equation}
r_{\text{oor}}=R_{\text{T}}^{\mathrm{aor}}\cdot\boldsymbol{1}_{\{e^{\mathrm{aor}}\}}+R_{\text{T}}^{\mathrm{gor}}\cdot\boldsymbol{1}_{\{e^{\mathrm{gor}}\}}
\end{equation}
where both transition rewards are negative penalties. The condition
$R_{\text{T}}^{\mathrm{aor}}=R_{\text{T}}^{T_{1}\rightarrow T_{3}}<R_{\text{T}}^{\mathrm{gor}}=R_{\text{T}}^{T_{2}\rightarrow T_{3}}$
reflects greater task progress when transitioning from the grasping
stage compared to the approach stage. The indicator function $\ensuremath{\boldsymbol{1}_{\{e\}}}$
equals 1 if the event $e$ occurs and 0 otherwise. It ensures the
right penalty is applied only when specific transitions occur. Here,
$\ensuremath{e^{\mathrm{aor}}}=\ensuremath{e^{T_{1}\rightarrow T_{3}}}$
and $\ensuremath{e^{\mathrm{gor}}}=\ensuremath{e^{T_{2}\rightarrow T_{3}}}$.

The reward function for the grasp-failure stage is 
\begin{equation}
r_{\text{fail}}=R_{\text{T}}^{\mathrm{fail}}
\end{equation}
where $R_{\text{T}}^{\mathrm{fail}}=R_{\text{T}}^{T_{2}\rightarrow T_{4}}$
is a smaller penalty compared to the out-of-reach stage, as it acknowledges
partial task completion.

In addition to monitoring task outcomes, the grasp-success stage evaluates
grasp quality through a friction cone analysis. The reward function
of this stage is defined as
\begin{equation}
r_{\text{succ}}=R_{\text{T}}^{\mathrm{succ}}+R_{\text{\text{cone}}}\cdot\boldsymbol{1}_{\{o_{\text{cone}}\}}
\end{equation}
where $R_{\text{T}}^{\mathrm{succ}}=R_{\text{T}}^{T_{2}\rightarrow T_{5}}$
is a large reward for successfully transitioning to this stage, and
$R_{\text{\text{cone}}}$ is an additional reward given only when
all contact points satisfy the friction cone condition. This reward
design encourages both task completion and a stable grasp. 

\section{Experiments}

The proposed method is evaluated in a simulated environment and compared
with the state-of-the-art methods. To validate its real-world applicability,
the method is transferred to a physical robot for performance testing.
The evaluation results are analyzed and discussed in this section.

\subsection{Experiment Setup}

\subsubsection{Environment Setup}

\begin{figure}
\centering{}\includegraphics[width=0.9\columnwidth]{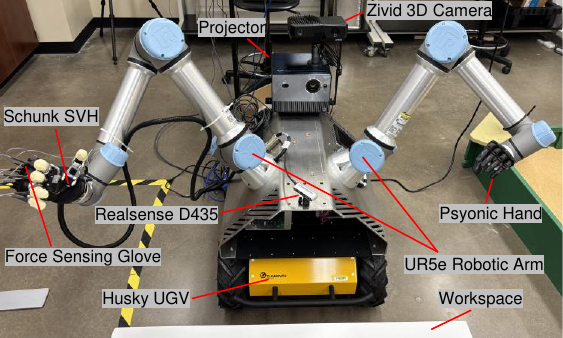}

\caption{Real-world experiment setup for grasping tasks. \label{fig: dual-arm-robot}}
\end{figure}

The task environment for the grasping task is illustrated in Fig.
\ref{fig: dual-arm-robot}. The target object is placed on a table
in front of a dual-arm mobile robot. This robot comprises a Husky
UGV (Unmanned Ground Vehicle) for mobility, two UR5e robotic arms,
a Schunk SVH robotic hand (right), and a PSYONIC Ability Hand (left).
Each robotic hand is mounted on a UR5e arm, and both arms are attached
to the Husky UGV. This configuration enables coordinated manipulation.

The robot integrates various sensors to enhance perception and interaction.
The Schunk SVH hand is equipped with an ErgoGLOVE Force Sensing System
for contact force detection, while the PSYONIC Ability Hand features
built-in force sensors. A RealSense D435 depth camera provides visual
input which enables precise object recognition, object pose estimation,
and environmental awareness. Additionally, a Zivid One 3D camera and
a projector are also part of the robot but are not utilized in this
study. 

A simulation environment was developed using PyBullet and OpenAI Gym
to train and test the proposed model. It replicates the real-world
setup of the robot, which enables seamless transfer of learned policies
to the physical robot for performance evaluation and practical applications.

\subsubsection{Dataset}

The AffordPose dataset serves as a benchmark for robotic grasping
tasks which emphasizes affordance-based pose estimation \cite{jian2023affordpose}.
It provides 3D object models, annotated grasp poses, and corresponding
affordance labels. 
\begin{figure*}[t]
\centering{}\includegraphics[width=0.96\textwidth]{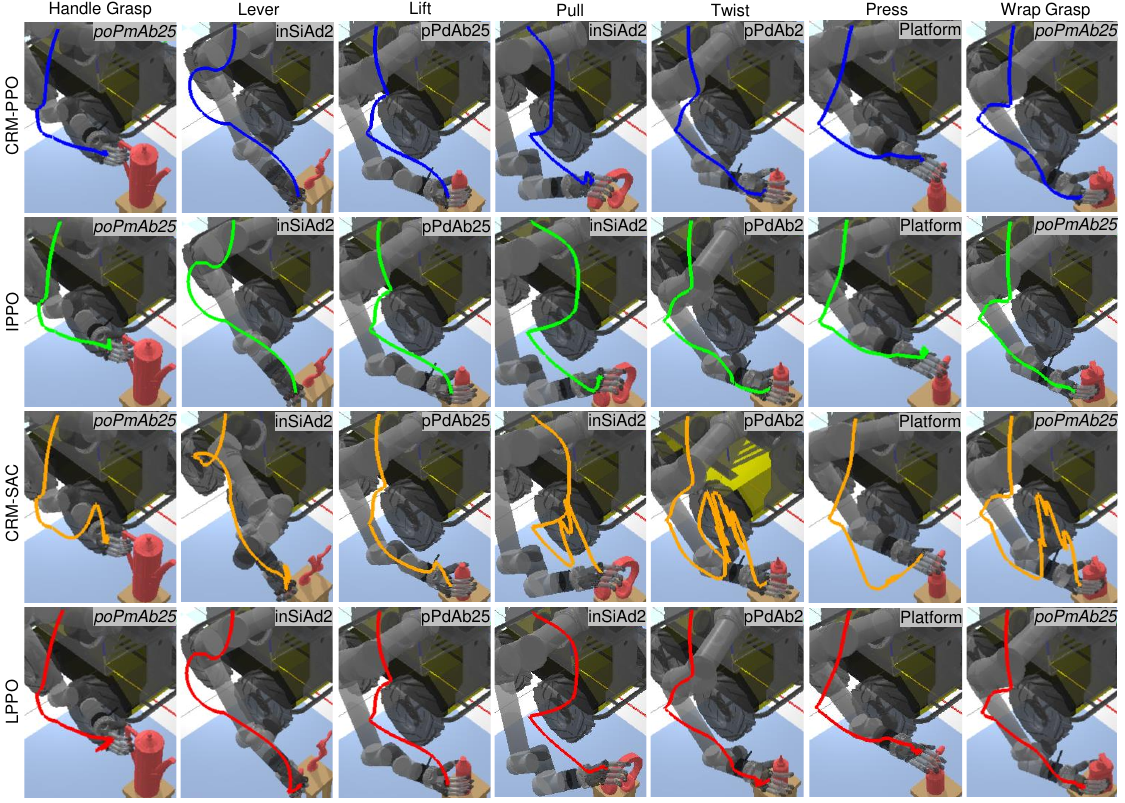}

\caption{Trajectory visualization of grasping tasks performed by the proposed
model and baseline models across different task objectives and grasp
topologies.\label{fig:  simu_exp}}
\end{figure*}

For this work, we refined the AffordPose dataset by removing redundant
entries, classifying grasp poses based on the grasp taxonomy defined
in Fig. \ref{fig: topology}, and computing grasp locations to support
the proposed grasping task framework. The revised dataset comprises
six unique grasp poses, seven distinct grasping objectives, and 20
diverse objects. As a result, the dataset contains a total of 26 different
grasping tasks across various objects, purposes, and grasp configurations. 

\subsection{Performance Evaluation in the Simulation Environment}

\subsubsection{Evaluation Metrics}

The proposed CRM-PPO model was evaluated against three baseline models
using the benchmark dataset. The evaluation is divided into two parts:
(1) performance assessment within the CRM framework and (2) comparison
with state-of-the-art methods.

The first baseline, CRM-SAC, integrates the CRM framework with the
Soft Actor-Critic (SAC) algorithm. This setup enables a direct comparison
with CRM-PPO under identical conditions, which demonstrates the superior
performance of PPO within the CRM framework. The second and third
baselines, LPPO (Learning-based PPO) and IPPO (Improved PPO), represent
state-of-the-art approaches for grasping tasks based on stage-wise
mechanisms. LPPO employs hierarchical dense rewards to enhance training
efficiency and generalization across diverse object configurations
\cite{shahid2020learning}. IPPO employs a stage-wise sparse reward
structure, which improves convergence speed and grasping accuracy
\cite{zhang2022simulation}. The comparison with these baselines confirms
the performance advantages introduced by the CRM framework.

The evaluation metrics include task success rate and average task
completion time in timesteps. 

\subsubsection{Training Setup}

The proposed model\footnote{This model is available at https://github.com/hhelium/DexMobile}
is trained for approximately 12,000 episodes (equivalent to 5 million
timesteps) using a discount factor of 0.99, a GAE lambda of 0.95,
and a batch size of 64. A dynamic learning rate is applied, starting
at $3\times10^{-5}$ for the first 40$\%$ of training progress. Between
40$\%$ and 70$\%$ progress, the learning rate is reduced to 90$\%$
of its initial value. During the final 30$\%$ of training, it is
further reduced to 80$\%$.

During training, the robot hand performs grasping tasks with different
task objectives and grasp topologies on a variety of objects. To simulate
real-world uncertainties, random noise of $\pm3\,\text{mm}$ in object
position, $\pm11.5^{\circ}$ in object orientation, and $0.02\,\text{rad}$
in joint positions was applied. This domain randomization method enhances
the model\textquoteright s robustness and generalization to real-world
scenarios by introducing randomized variations in object pose and
joint configurations during training.

For task objectives such as handle grasp, lift, lever, pull, and wrap
grasp, a task is considered successful if the robot picks up the object
and holds it steadily for 5 seconds without dropping it. For the twist
objective, success is defined as applying sufficient torque in the
twisting direction after grasping the object. In the press objective,
success is achieved if the robot applies enough force in the pressing
direction. 
\begin{table*}[t]
\caption{The testing results for the proposed and baseline models. \label{tab:The-testing-resuts}}

\centering{}%
\begin{tabular}{lllllllll}
\toprule 
\multirow{3}{*}{Models} & \multicolumn{2}{c}{\textbf{IPPO }\cite{zhang2022simulation}} & \multicolumn{2}{c}{\textbf{LPPO }\cite{shahid2020learning}} & \multicolumn{2}{c}{\textbf{CRM-SAC}} & \multicolumn{2}{c}{\textbf{CRM-PPO (this paper)}}\tabularnewline
 & Success & Episode & Success & Episode & Success & Episode & Success & Episode\tabularnewline
 & Rate & Length & Rate & Length & Rate & Length & Rate & Length\tabularnewline
\midrule
\textbf{Lift} & 0.85 & 412.63 & 0.88 & 456.33 & 0.63 & 534.68 & \textbf{0.88} & \textbf{342.43}\tabularnewline
\textbf{Pull} & 0.86 & 433.60 & 0.76 & 315.82 & \textbf{1.0} & 276.22 & \textbf{1.0} & \textbf{217.34}\tabularnewline
\textbf{Press} & 0.76 & 415.82 & 0.29 & 793.22 & 0.33 & 787.64 & \textbf{0.96} & \textbf{167.30}\tabularnewline
\textbf{Twist} & 0.74 & 436.20 & 0.67 & 531.34 & 0.49 & 599.17 & \textbf{0.91} & \textbf{335.52}\tabularnewline
\textbf{Lever} & 0.70 & 466.80 & 0.90 & 367.67 & \textbf{0.95} & 307.24 & 0.93 & \textbf{246.15}\tabularnewline
\textbf{Wrap-Grasp} & 0.89 & 348.88 & 0.67 & 372.49 & 0.60 & 377.41 & \textbf{1.0} & \textbf{252.17}\tabularnewline
\textbf{Handle-Grasp} & \textbf{0.98} & 303.70 & 0.97 & 369.71 & 0.75 & 338.43 & 0.97 & \textbf{281.09}\tabularnewline
\textbf{Overall} & 0.84 & 390.86 & 0.71 & 461.75 & 0.61 & 479.36 & \textbf{0.95} & \textbf{273.07}\tabularnewline
\bottomrule
\end{tabular}
\end{table*}

An early stopping mechanism is implemented to improve training efficiency.
The training process is terminated when the average success rate of
the most recent 100 episodes reaches or exceeds 99$\%$. 

Examples of trajectories generated by the proposed and baseline methods
are shown in Fig. \ref{fig:  simu_exp}. The figure illustrates that
the trajectory generated by the PPO-based method is smoother and more
optimized compared to that of the SAC-based method. This difference
may be due to SAC's lower efficiency in tasks with well-shaped rewards.

For the PPO-based method, both the proposed and baseline approaches
exhibit similar trajectories. The proposed method completes the trajectory
significantly faster and avoids overlapping or revisiting previous
paths. This improvement is due to the stage-specific context and transition
rewards, which effectively guide the model to complete the task efficiently
and without redundancy. 

\subsubsection{Result Analysis}

\begin{figure}[h]
\centering{}\includegraphics[width=0.98\columnwidth]{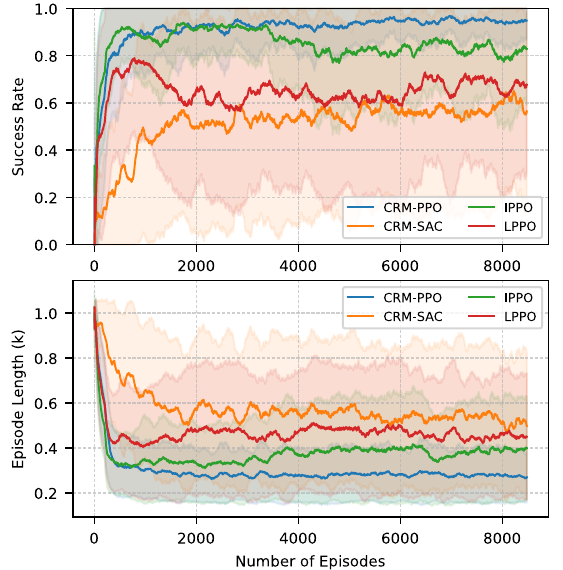}

\caption{Comparison of different models in grasping tasks based on success
rate and episode length. \label{fig:  traning_performance}}
\end{figure}

The proposed CRM-PPO model and the three baselines were trained five
times each. For each model, the average success rate and average episode
length were calculated as evaluation metrics, along with their standard
deviations to reflect performance variability across runs. The training
results are presented in Fig.\,\ref{fig:  traning_performance}, where
solid curves represent average values and shaded areas indicate standard
deviations. 

The proposed CRM-PPO model outperforms all baselines across all evaluation
metrics. It records the highest average success rate and the shortest
average episode length. The model reaches the early termination criterion
after approximately 8,200 episodes on average, while none of the baseline
models meet this criterion within the full training period. All five
runs of CRM-PPO reach early termination and show the lowest standard
deviation. These results reflect superior robustness, consistency,
and sample efficiency. This strong performance results from its hierarchical
reward structure, which provides clear stage-specific guidance and
supports effective transitions between stages through transition rewards.

The performance of IPPO surpasses that of LPPO, primarily due to differences
in reward design. IPPO assigns sparse rewards based on stage transitions,
while LPPO defines continuous intra-stage rewards without explicitly
encouraging transitions. In summary, IPPO provides only transition
rewards, whereas LPPO offers only intra-stage rewards. IPPO outperforms
LPPO because its sparse reward structure effectively guides the agent
toward target stages. In contrast, LPPO focuses more on intra-stage
optimization and lacks incentives for transitioning between stages,
which results in lower overall performance despite its intra-stage
guidance.

During early training, IPPO\textquoteright s performance approximates
that of the proposed CRM-PPO model, as its sparse reward structure
provides implicit transition-based guidance. Due to the absence of
intra-stage rewards, IPPO does not support fine-grained exploration
and adaptation, which reduces data efficiency. As training progresses,
this limitation causes IPPO's performance to lag behind CRM-PPO.

The CRM-PPO model effectively integrates both approaches by combining
stage-specific guidance and transition rewards, and it outperforms
the baseline models across evaluation metrics. CRM-SAC performs the
worst across all evaluation metrics, which indicates that PPO-based
methods are better suited for hierarchical grasping tasks. 

The proposed model and baseline models were evaluated using a benchmark
dataset, with each model tested 1,000 times on randomly selected tasks.
The results are summarized in Table \ref{tab:The-testing-resuts}.
 The proposed model consistently outperformed baseline models in
most tasks. Notably, it excelled in the challenging Twist task, where
baseline models showed relatively low success rates. This task demands
precise and coordinated actions, it demonstrates CRM-PPO's ability
to handle complex scenarios due to its task-specific structured design,
which provides effective execution guidance.

In the Lever and Handle-Grasp tasks, the proposed model achieved slightly
lower success rates, trailing the best-performing baseline by 2$\%$
and 1$\%$, respectively. Despite this, it surpassed all baselines
in episode length and completed tasks more efficiently. These results
confirm the effectiveness of the CRM-PPO model in task-oriented grasping
tasks.

\subsection{Real-World Evaluation}

The robots are integrated through ROS2 Humble on Ubuntu 22.04 with
a low-latency kernel to support real-time performance. Inverse kinematics
and collision avoidance are managed through MoveIt2. An Intel RealSense
D435 depth camera is spatially aligned with the robot's coordinate
frame via an eye-to-hand calibration, which enables accurate perception
and interaction within the shared workspace. The depth images align
with the color images, both with a resolution of 640 \texttimes{}
480 pixels. The image streams and force feedback data from the ErgoGLOVE
Force Sensing System are synchronized and published to the proposed
model through ROS2 topics. This setup ensures coherent sensory input
for perception and interaction tasks.

Even though the simulation and the real robot are identical in design,
a sim-to-real gap persists due to discrepancies in dynamic properties,
environmental conditions, and sensor noise. To bridge this gap, domain
randomization is employed \cite{tobin2017domain}. The policy is trained
across numerous variations of simulation parameters, such as joint
position error, object pose error, and sensor noise. By randomizing
these parameters over a wide range, the likelihood of the policy generalizing
to real-world conditions increases significantly.

The robot perceives the object's pose and the contact forces between
the Schunk hand and the object. The object pose is estimated using
FoundationPose \cite{wen2024foundationpose}, which provides millimeter-level
accuracy in pose estimation. Based on the object pose, the observations
$o_{\textnormal{dist}}$, $\boldsymbol{o}_{\textnormal{object}}$,
and $\boldsymbol{o}_{\mathrm{\text{relative}}}$ can be calculated.
The ErgoGLOVE Force Sensing System detects the contact force between
the Schunk hand and the object and provides $n_{\text{c}}$. Since
the force-sensing glove measures force along only one axis, $\boldsymbol{o}_{\mathrm{\text{force}}}$,
$\boldsymbol{o}_{\mathrm{\text{torque}}}$, and $o_{\textnormal{cone}}$
cannot be detected and are set to zero. 

In the simulation environment, the grasping force cannot be determined
properly due to the agnostic nature of the target object's material
properties, as it is represented by a 3D model. As a result, the grasping
force cannot be accurately calibrated, and the object remains unaffected
by any applied force. In contrast, in the real world, the contact
force must be carefully controlled to avoid damaging the object or
the robotic hand. To address this discrepancy and enable the transfer
of the simulation model to a physical robot, we manually defined grasping
force thresholds. For fragile objects, the contact force was constrained
to a maximum of $1\,\text{N}$, with the robotic fingers halting motion
upon reaching this threshold. For sturdy objects, the threshold was
set between $1\,\text{N}$ and $3\,\text{N}$, depending on factors
such as the object's texture and weight. 
\begin{figure}[h]
\centering{}\includegraphics[width=0.96\columnwidth]{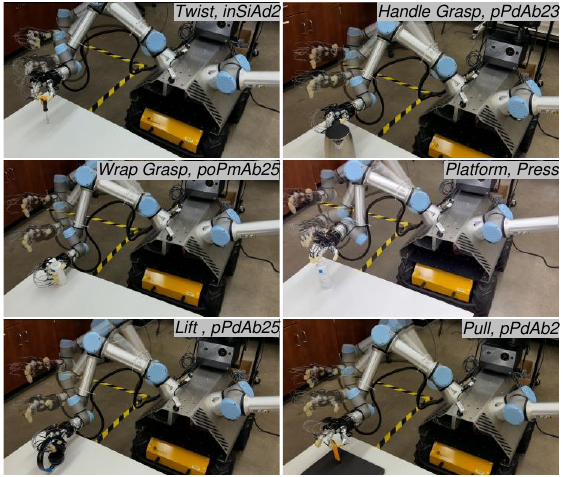}

\caption{Real-world grasping tasks across different affordances and grasp topologies.
\label{fig: real-grasp}}
\end{figure}

The proposed CRM-PPO model is fine-tuned for 300 episodes across six
affordances, including Handle Grasp, Lift, Press, Pull, Twist, and
Wrap Grasp, each utilizing the corresponding grasp topology, as shown
in Fig. \ref{fig: real-grasp}. The achieved testing success rates
for these tasks are 100\%, 90\%, 60\%, 80\%, 70\%, and 100\%, respectively.
In the experiment, the grasping tasks for press and twist exhibited
lower success rates of 60\% and 70\%, respectively. The low success
rate for the twist task is attributed to the small, round-shaped handle
of the screwdriver, which is prone to slipping when grasped using
the topology inSiAd2. Additionally, the screwdriver, when placed on
the table, can be easily knocked over with even light contact from
the hand. The press task had the lowest success rate due to the limitations
of the sensing glove. The eight force sensors on the glove cover only
a limited area of the hand, and during pressing, the actual contact
area often lacks sensor coverage, which can result in task failure.
After sensor repositioning, the success rate improved significantly,
while success rates for other tasks decreased. The overall success
rate achieved was 83.3\%, which can be further improved by adding
more force sensors and further fine-tuning.

\section{Conclusion}

This paper presents a context-aware task-oriented dexterous grasping
approach leveraging a Contextual Reward Machine framework. The CRM
decomposes complex grasping tasks into modular sub-tasks with stage-specific
contexts, which enables efficient learning and execution. By integrating
Proximal Policy Optimization, the proposed method achieves significant
improvements in learning efficiency, task performance, and adaptability.
Extensive experiments in simulated and real-world environments validated
the effectiveness and robustness of the proposed approach. The CRM-PPO
model achieved a 95\% success rate in simulation across 1,000 grasping
tasks. When transferred to a real robot, it attained an 83.3\% success
rate over 60 real-world tasks. The proposed model exceeded the performance
of state-of-the-art models in success rate and task completion time.
Its ability to adapt to diverse grasping objectives, various grasp
topologies, and dynamic conditions highlights its practical applicability
in unstructured environments. The results demonstrate the potential
of the CRM-PPO framework to advance robotic dexterity and manipulation. 

\bibliographystyle{IEEEtran}
\bibliography{reff}

@Article{bullock2015yale,
  Title                    = {The Yale human grasping dataset: Grasp, object, and task data in household and machine shop environments},
  Author                   = {Bullock, Ian M and Feix, Thomas and Dollar, Aaron M},
  Journal                  = {The International Journal of Robotics Research},
  Year                     = {2015},
  Number                   = {3},
  Pages                    = {251--255},
  Volume                   = {34},

  Publisher                = {SAGE Publications Sage UK: London, England}
}

@Article{cutkosky1989grasp,
  Title                    = {On grasp choice, grasp models, and the design of hands for manufacturing tasks.},
  Author                   = {Cutkosky, Mark R and others},
  Journal                  = {IEEE Transactions on robotics and automation},
  Year                     = {1989},
  Number                   = {3},
  Pages                    = {269--279},
  Volume                   = {5}
}

@Article{deimel2016novel,
  Title                    = {A novel type of compliant and underactuated robotic hand for dexterous grasping},
  Author                   = {Deimel, Raphael and Brock, Oliver},
  Journal                  = {The International Journal of Robotics Research},
  Year                     = {2016},
  Number                   = {1-3},
  Pages                    = {161--185},
  Volume                   = {35},

  Publisher                = {SAGE Publications Sage UK: London, England}
}

@Article{feix2014analysis,
  Title                    = {Analysis of human grasping behavior: Correlating tasks, objects and grasps},
  Author                   = {Feix, Thomas and Bullock, Ian M and Dollar, Aaron M},
  Journal                  = {IEEE transactions on haptics},
  Year                     = {2014},
  Number                   = {4},
  Pages                    = {430--441},
  Volume                   = {7},

  Publisher                = {IEEE}
}

@Article{feix2016grasp,
  Title                    = {The grasp taxonomy of human grasp types},
  Author                   = {Feix, Thomas and Romero, Javier and Schmiedmayer, Heinz-Bodo and Dollar, Aaron M and Kragic, Danica},
  Journal                  = {IEEE Transactions on Human-Machine Systems},
  Year                     = {2016},
  Number                   = {1},
  Pages                    = {66--77},
  Volume                   = {46},

  Publisher                = {IEEE}
}

@InProceedings{li2021learning,
  Title                    = {Learning Task-Oriented Dexterous Grasping from Human Knowledge},
  Author                   = {Li, Hui and Zhang, Yinlong and Li, Yanan and He, Hongsheng},
  Booktitle                = {2021 IEEE International Conference on Robotics and Automation (ICRA)},
  Year                     = {2021},
  Organization             = {IEEE},
  Pages                    = {6192--6198}
}

@InProceedings{rao2019object,
  Title                    = {Object recall from natural-language descriptions for autonomous robotic grasping},
  Author                   = {Rao, Achyutha Bharath and Li, Hui and He, Hongsheng},
  Booktitle                = {2019 IEEE International Conference on Robotics and Biomimetics (ROBIO)},
  Year                     = {2019},
  Organization             = {IEEE},
  Pages                    = {1368--1373}
}

@Article{santello1998postural,
  Title                    = {Postural hand synergies for tool use},
  Author                   = {Santello, Marco and Flanders, Martha and Soechting, John F},
  Journal                  = {Journal of Neuroscience},
  Year                     = {1998},
  Number                   = {23},
  Pages                    = {10105--10115},
  Volume                   = {18},

  Publisher                = {Soc Neuroscience}
}

@Article{yang2019deep,
  author  = {Yang, Yuguang},
  title   = {A Deep Reinforcement Learning Architecture for Multi-stage Optimal Control},
  journal = {arXiv preprint arXiv:1911.10684},
  year    = {2019},
}

@Article{james2020slip,
  author    = {James, Jasper Wollaston and Lepora, Nathan F},
  title     = {Slip detection for grasp stabilization with a multifingered tactile robot hand},
  journal   = {IEEE Transactions on Robotics},
  year      = {2020},
  volume    = {37},
  number    = {2},
  pages     = {506--519},
  publisher = {IEEE},
}

@Article{thuruthel2018model,
  author    = {Thuruthel, Thomas George and Falotico, Egidio and Renda, Federico and Laschi, Cecilia},
  title     = {Model-based reinforcement learning for closed-loop dynamic control of soft robotic manipulators},
  journal   = {IEEE Transactions on Robotics},
  year      = {2018},
  volume    = {35},
  number    = {1},
  pages     = {124--134},
  publisher = {IEEE},
}

@Article{stulp2012reinforcement,
  author    = {Stulp, Freek and Theodorou, Evangelos A and Schaal, Stefan},
  title     = {Reinforcement learning with sequences of motion primitives for robust manipulation},
  journal   = {IEEE Transactions on robotics},
  year      = {2012},
  volume    = {28},
  number    = {6},
  pages     = {1360--1370},
  publisher = {IEEE},
}

@Article{he2020reinforcement,
  author    = {He, Wei and Gao, Hejia and Zhou, Chen and Yang, Chenguang and Li, Zhijun},
  title     = {Reinforcement learning control of a flexible two-link manipulator: an experimental investigation},
  journal   = {IEEE Transactions on Systems, Man, and Cybernetics: Systems},
  year      = {2020},
  volume    = {51},
  number    = {12},
  pages     = {7326--7336},
  publisher = {IEEE},
}

@Article{mentzel2011dynamics,
  author  = {Mentzel, M and Benlic, A and Wachter, NJ and Gulkin, D and Bauknecht, S and G{\"u}lke, J},
  title   = {The dynamics of motion sequences of the finger joints during fist closure},
  journal = {Handchirurgie, Mikrochirurgie, Plastische Chirurgie: Organ der Deutschsprachigen Arbeitsgemeinschaft fur Handchirurgie: Organ der Deutschsprachigen Arbeitsgemeinschaft fur Mikrochirurgie der Peripheren Nerven und Gefasse: Organ der V...},
  year    = {2011},
  volume  = {43},
  number  = {3},
  pages   = {147--154},
}

@Article{roda2022studying,
  author    = {Roda-Sales, Alba and Sancho-Bru, Joaqu{\'\i}n L and Vergara, Margarita},
  title     = {Studying kinematic linkage of finger joints: estimation of kinematics of distal interphalangeal joints during manipulation},
  journal   = {PeerJ},
  year      = {2022},
  volume    = {10},
  pages     = {e14051},
  publisher = {PeerJ Inc.},
}

@InProceedings{hrabia2013whole,
  author    = {Hrabia, Christopher-Eyk and Wolf, Katrin and Wilhelm, Mathias},
  title     = {Whole hand modeling using 8 wearable sensors: Biomechanics for hand pose prediction},
  booktitle = {Proceedings of the 4th Augmented Human International Conference},
  year      = {2013},
  pages     = {21--28},
}

@Article{sutton2018reinforcement,
  author  = {Sutton, Richard S},
  title   = {Reinforcement learning: An introduction},
  journal = {A Bradford Book},
  year    = {2018},
}

@Article{icarte2022reward,
  author  = {Icarte, Rodrigo Toro and Klassen, Toryn Q and Valenzano, Richard and McIlraith, Sheila A},
  journal = {Journal of Artificial Intelligence Research},
  title   = {Reward machines: Exploiting reward function structure in reinforcement learning},
  year    = {2022},
  pages   = {173--208},
  volume  = {73},
}

@InProceedings{jian2023affordpose,
  author    = {Jian, Juntao and Liu, Xiuping and Li, Manyi and Hu, Ruizhen and Liu, Jian},
  booktitle = {Proceedings of the IEEE/CVF International Conference on Computer Vision},
  title     = {Affordpose: A large-scale dataset of hand-object interactions with affordance-driven hand pose},
  year      = {2023},
  pages     = {14713--14724},
}

@InProceedings{zhang2022simulation,
  author       = {Zhang, Zhizhuo},
  booktitle    = {Journal of Physics: Conference Series},
  title        = {Simulation of robotic arm grasping control based on proximal policy optimization algorithm},
  year         = {2022},
  number       = {1},
  organization = {IOP Publishing},
  pages        = {012065},
  volume       = {2203},
}

@InProceedings{shahid2020learning,
  author       = {Shahid, Asad Ali and Roveda, Loris and Piga, Dario and Braghin, Francesco},
  booktitle    = {2020 IEEE International Conference on Systems, Man, and Cybernetics (SMC)},
  title        = {Learning continuous control actions for robotic grasping with reinforcement learning},
  year         = {2020},
  organization = {IEEE},
  pages        = {4066--4072},
}

@Article{schulman2017proximal,
  author  = {Schulman, John and Wolski, Filip and Dhariwal, Prafulla and Radford, Alec and Klimov, Oleg},
  journal = {arXiv preprint arXiv:1707.06347},
  title   = {Proximal policy optimization algorithms},
  year    = {2017},
}

@InProceedings{fan2017robust,
  author       = {Fan, Yongxiang and Sun, Liting and Zheng, Minghui and Gao, Wei and Tomizuka, Masayoshi},
  booktitle    = {2017 IEEE International Conference on Advanced Intelligent Mechatronics (AIM)},
  title        = {Robust dexterous manipulation under object dynamics uncertainties},
  year         = {2017},
  organization = {IEEE},
  pages        = {613--619},
}

@InProceedings{wang2023dexgraspnet,
  author       = {Wang, Ruicheng and Zhang, Jialiang and Chen, Jiayi and Xu, Yinzhen and Li, Puhao and Liu, Tengyu and Wang, He},
  booktitle    = {2023 IEEE International Conference on Robotics and Automation (ICRA)},
  title        = {Dexgraspnet: A large-scale robotic dexterous grasp dataset for general objects based on simulation},
  year         = {2023},
  organization = {IEEE},
  pages        = {11359--11366},
}

@InProceedings{mandikal2022dexvip,
  author       = {Mandikal, Priyanka and Grauman, Kristen},
  booktitle    = {Conference on Robot Learning},
  title        = {Dexvip: Learning dexterous grasping with human hand pose priors from video},
  year         = {2022},
  organization = {PMLR},
  pages        = {651--661},
}

@Article{duan2021robotics,
  author    = {Duan, Haonan and Wang, Peng and Huang, Yayu and Xu, Guangyun and Wei, Wei and Shen, Xiaofei},
  journal   = {Frontiers in Neurorobotics},
  title     = {Robotics dexterous grasping: The methods based on point cloud and deep learning},
  year      = {2021},
  pages     = {658280},
  volume    = {15},
  publisher = {Frontiers Media SA},
}

@Article{shang2020deep,
  author    = {Shang, Weiwei and Song, Fangjing and Zhao, Zengzhi and Gao, Hongbo and Cong, Shuang and Li, Zhijun},
  journal   = {IEEE Transactions on Cybernetics},
  title     = {Deep learning method for grasping novel objects using dexterous hands},
  year      = {2020},
  number    = {5},
  pages     = {2750--2762},
  volume    = {52},
  publisher = {IEEE},
}

@InProceedings{qin2023dexpoint,
  author       = {Qin, Yuzhe and Huang, Binghao and Yin, Zhao-Heng and Su, Hao and Wang, Xiaolong},
  booktitle    = {Conference on Robot Learning},
  title        = {Dexpoint: Generalizable point cloud reinforcement learning for sim-to-real dexterous manipulation},
  year         = {2023},
  organization = {PMLR},
  pages        = {594--605},
}

@InProceedings{mandikal2021learning,
  author       = {Mandikal, Priyanka and Grauman, Kristen},
  booktitle    = {2021 IEEE international conference on robotics and automation (ICRA)},
  title        = {Learning dexterous grasping with object-centric visual affordances},
  year         = {2021},
  organization = {IEEE},
  pages        = {6169--6176},
}

@InProceedings{tobin2017domain,
  author       = {Tobin, Josh and Fong, Rachel and Ray, Alex and Schneider, Jonas and Zaremba, Wojciech and Abbeel, Pieter},
  booktitle    = {2017 IEEE/RSJ international conference on intelligent robots and systems (IROS)},
  title        = {Domain randomization for transferring deep neural networks from simulation to the real world},
  year         = {2017},
  organization = {IEEE},
  pages        = {23--30},
}

@InProceedings{wen2024foundationpose,
  author    = {Wen, Bowen and Yang, Wei and Kautz, Jan and Birchfield, Stan},
  booktitle = {Proceedings of the IEEE/CVF Conference on Computer Vision and Pattern Recognition},
  title     = {Foundationpose: Unified 6d pose estimation and tracking of novel objects},
  year      = {2024},
  pages     = {17868--17879},
}

@Article{chen2025target,
  author    = {Chen, Yun and Xiao, Jiaping},
  journal   = {Machine Intelligence Research},
  title     = {Target search and navigation in heterogeneous robot systems with deep reinforcement learning},
  year      = {2025},
  number    = {1},
  pages     = {79--90},
  volume    = {22},
  publisher = {Springer},
}

@InProceedings{chen2025towards,
  author       = {Chen, Yun and Zhang, Xinyu and Li, Hui and He, Hongsheng and Shou, Wan and Zhang, Qiang},
  booktitle    = {2025 IEEE International Conference on Robotics and Automation (ICRA)},
  title        = {Towards Neurorobotic Interface for Finger Joint Angle Estimation: A Multi-Stage CNN-LSTM Network with Transfer Learning},
  year         = {2025},
  organization = {IEEE},
  pages        = {4409--4416},
}

@Article{li2022reinforcement,
  author    = {Li, Long and Luo, Yu and Yang, Jing and Pu, Lina},
  journal   = {IEEE Transactions on Information Forensics and Security},
  title     = {Reinforcement learning enabled intelligent energy attack in green IoT networks},
  year      = {2022},
  pages     = {644--658},
  volume    = {17},
  publisher = {IEEE},
}

@InProceedings{li2021q,
  author       = {Li, Long and Luo, Yu and Pu, Lina},
  booktitle    = {ICC 2021-IEEE International Conference on Communications},
  title        = {Q-learning enabled intelligent energy attack in sustainable wireless communication networks},
  year         = {2021},
  organization = {IEEE},
  pages        = {1--6},
}

\end{document}